\documentclass[10pt]{article}

\usepackage[preprint]{tmlr}

\usepackage{microtype}
\usepackage{graphicx}
\usepackage{longtable}
\usepackage{multirow}
\usepackage{subcaption}
\usepackage{booktabs} % for professional tables
\usepackage[usenames, dvipsnames]{xcolor}
\usepackage{hyperref}
\usepackage{url}
\usepackage{xspace}

\newcommand{\Noptimizers}{15\xspace}
\newcommand{\Ndatasets}{17\xspace}

\usepackage{amsmath}
\usepackage{amssymb}
\usepackage{mathtools}
\usepackage{amsthm}

\usepackage{enumitem}
\setlist[itemize]{leftmargin=*,topsep=2pt,itemsep=2pt,parsep=0pt,partopsep=0pt}

\usepackage[capitalize,noabbrev]{cleveref}

\theoremstyle{plain}

\theoremstyle{definition}

\theoremstyle{remark}

\usepackage[textsize=tiny]{todonotes}
\usepackage{pifont}

\usepackage[usenames, dvipsnames]{xcolor}

\definecolor{TMLRLinkBlue}{HTML}{1D3557} % deep academic blue
\definecolor{purple}{HTML}{7B2D8E}
\definecolor{darkpurple}{HTML}{82427F}
\definecolor{blue}{HTML}{2b527c}

\hypersetup{
    colorlinks=true,
    linkcolor=TMLRLinkBlue,
    citecolor=TMLRLinkBlue,
    urlcolor=TMLRLinkBlue,
    filecolor=TMLRLinkBlue,
    pdfborder={0 0 0}
}

\usepackage{tcolorbox}

\title{\centering Benchmarking Optimizers \\ for MLPs in Tabular Deep Learning}
\author{
\begin{center}
\name Yury Gorishniy$^{\alpha,\mathbf{=}}$  \name Ivan Rubachev$^{\alpha,\beta,\mathbf{=}}$ Dmitrii Feoktistov$^{\beta,\alpha}$   \name Artem Babenko$^{\alpha,\beta}$
\\[0.6em]
{\normalsize\normalfont $^{\alpha}$ Yandex \qquad $^{\beta}$ HSE University}
\end{center}
}

\def\month{03}
\def\year{2026}
\def\openreview{\url{https://openreview.net/forum?id=XXXX}}

\begin{document}

\maketitle

\vspace{-2em}

\begin{abstract}

\vspace{-0.7em}
MLP is a heavily used backbone in modern deep learning (DL) architectures for supervised learning on tabular data, and AdamW is the go-to optimizer used to train tabular DL models.
Unlike architecture design, however, the choice of optimizer for tabular DL has not been examined systematically, despite new optimizers showing promise in other domains.
To fill this gap, we benchmark \Noptimizers optimizers on \Ndatasets tabular datasets for training MLP-based models in the standard supervised learning setting under a shared experiment protocol.
Our main finding is that the Muon optimizer consistently outperforms AdamW, and thus should be considered a strong and practical choice for practitioners and researchers, if the associated training efficiency overhead is affordable.
Additionally, we find exponential moving average of model weights to be a simple yet effective technique that improves AdamW on vanilla MLPs, though its effect is less consistent across model variants.
The code is available at \url{https://github.com/yandex-research/tabular-dl-optimizers}.

\end{abstract}

\vspace{-1em}
\section{Introduction}

In supervised learning on tabular data, multi-layer perceptrons (MLPs) and MLP-based architectures remain widely used and practically important. MLPs are not only the standard baseline, but also the foundation of many of the strongest recent methods \citep{gorishniy2022embeddings, gorishniy2024tabr, gorishniy2025tabm, ye2025revisiting, holzmuller2024better}. This makes their training recipe an important practical question. Yet while tabular deep learning has seen extensive work on architectures, optimization has remained comparatively under-examined, with AdamW \citep{loshchilov2018decoupled} serving as the de facto default.

Recent progress in optimization for deep learning makes this default worth revisiting. Muon \citep{jordan2024muon} has recently emerged as a strong new optimizer with promising empirical results across several domains ranging from LLM training \citep{modded_nanogpt_2024, liu2025muon} to Implicit Neural Representations \citep{mcginnis2025optimizing} and Information Retrieval \citep{takehi2025fantastic}. Furthermore, the recent AlgoPerf benchmark \citep{kasimbeg2025accelerating} showed that carefully tuned alternatives can outperform strong AdamW baselines across multiple training workloads. Together, these results suggest that AdamW may no longer be the best optimizer by default. 

Whether that conclusion extends to tabular deep learning, however, is not obvious. Tabular supervised learning often operates in noisy, finite-data regimes; training often relies on early stopping, and the end goal is generalization on held-out data -- not reaching a target loss faster. This means results from other domains may not transfer directly to
% standard
supervised tabular deep learning, which warrants an independent, tabular-specific benchmark.

\textbf{The main contributions} of our report are as follows:

\begin{enumerate}[nosep,leftmargin=2em]
    \item
    We benchmark \Noptimizers optimizers on \Ndatasets datasets for training MLP-based models for supervised learning tasks on tabular data, under a unified hyperparameter tuning and evaluation protocol.

    \item
    We find the Muon optimizer \citep{jordan2024muon} to consistently outperform AdamW for both plain MLPs and modern MLP-based architectures.
    The better performance of Muon comes at the cost of slower training; for stronger models, the efficiency overhead is less pronounced.

    \item
    We also highlight exponential moving average (EMA) of model weights as a simple way to improve AdamW \citep{loshchilov2018decoupled} for vanilla tabular MLPs.
    For more advanced tabular MLP-based models, however, the effect of EMA is rather less consistent.

\end{enumerate}

\section{Related Work}

\textbf{Deep Learning for Tabular Data}.
Current work on supervised learning on tabular data broadly follows two directions.
One focuses on the conventional
% in-weight
learning paradigm, where a model of a given architecture is trained from a random initialization on a target dataset \citep{gorishniy2025tabm,holzmuller2024better,ye2025revisiting}.
The other direction studies in-context-learning-based foundation models \citep{grinsztajn2025tabpfn,qu2026tabiclv2}.
In this report, we focus on the first direction.

\textbf{MLPs in Tabular Deep Learning}.
As multiple recent benchmarks show, the best-performing modern tabular DL architectures are those based on a multi-layer perceptron (MLP) and its variations \citep{zabergja2024tabular, erickson2025tabarena, rubachev2025tabred}.
Much of the recent progress has come from improving MLP-based models through architectural improvements and regularization techniques \citep{gorishniy2022embeddings, gorishniy2025tabm, holzmuller2024better}.
The choice of an optimizer for training such models, however, has not been systematically studied, and most studies use AdamW \citep{loshchilov2018decoupled} as the go-to option.

\textbf{Progress in Optimization for Deep Learning}. Several alternatives to AdamW have been proposed in recent years. Some remain close to the Adam family, modifying parts of the momentum and adaptive update rules \citep{Liu2020On, pagliardini2025the, taniguchi2024adopt, xie2024adan, dozat2016incorporating}. Others explore different update rules, including sign-based methods such as Lion \citep{chen2023symbolic} and Signum \citep{bernstein2018signsgd} and structured preconditioning methods such as Shampoo \citep{gupta2018shampoo}  or SOAP \citep{vyas2025soap}. Most notably, Muon \citep{jordan2024muon} has recently emerged as a strong optimizer, with promising empirical results reported in settings including language modeling \citep{modded_nanogpt_2024, liu2025muon}, implicit neural representations \citep{mcginnis2025optimizing}, information retrieval \citep{takehi2025fantastic}, reinforcement learning \citep{suarez2025pufferlib}, and most recently tabular foundation model pretraining \citep{qu2026tabiclv2}. These developments make optimizer choice a live practical question for tabular MLPs.

\textbf{Optimization-adjacent Techniques}. Beyond the choice of optimizer itself, several techniques have also proved useful in other domains. Two prominent examples are learning rate scheduling and weight averaging. In tabular deep learning, however, schedules are less convenient because training usually relies on early stopping rather than a fixed training budget. The Schedule-Free approach \citep{defazio2024road} is therefore especially relevant, since it aims to recover some of the benefits of scheduled training without requiring a predefined training horizon. Weight averaging techniques such as EMA and SWA have been used successfully in areas including computer vision \citep{izmailov2019averagingweightsleadswider, morales-brotons2024exponential}, generative modeling \citep{karras2024analyzing}, and large language model training \citep{li2025model}. The utility of these techniques in training MLP-based models on tabular data remains unclear.

\textbf{Benchmarking Optimizers}. Fairly comparing optimizers is difficult because results depend on many details of the experimental setup \citep{choi2020empiricalcomparisonsoptimizersdeep}. Recent efforts have made such comparisons more careful and standardized, including broad optimizer benchmarks such as AlgoPerf \citep{kasimbeg2025accelerating}, benchmarks specifically focused on optimizers for LLM pretraining \citep{semenov2025benchmarking}, or speedrun-style evaluations such as the NanoGPT speedrun \citep{modded_nanogpt_2024} -- a competition-style evaluation setup where the Muon optimizer was evaluated and presented early on. 

Importantly, the aforementioned benchmarks operate in a setup that differs from that of supervised tabular learning.
\textit{First}, those non-tabular benchmarks usually report the performance on the validation set as the final task metric.
At the same time, in tabular DL, models are often trained with early stopping based on the \textit{validation}-set performance, but the final metric is computed on a separate held-out \textit{test} set, which emphasizes the focus on generalization.
\textit{Second}, in non-tabular studies, the speed of reaching the target validation performance is often of interest.
In many tabular DL studies, as well as in this report, we focus more on the generalization rather than on efficiency.
\textit{Third,} tabular datasets are often noisier than those in other DL workloads \citep{grinsztajn2022why,kartashev2025unveiling}, which could also present a different set of challenges for optimization.

Overall, all these differences make supervised tabular deep learning a distinct empirical setting for optimizer evaluation.

\section{Experiments}
\label{sec:experiments}

In this section, we evaluate how optimizer choice affects the generalization performance of MLP-based models in supervised tabular learning under a unified experimental protocol.
We structure our experiments as follows.
First, we provide a complete overview of our tuning and evaluation setup (\S\ref{sec:setup}).
We then present an extensive empirical evaluation of a wide range of optimizers and optimization techniques for the MLP baseline (\S\ref{sec:main-benchmark}).
Then, we evaluate the most performant and effective optimizer variants for a broader set of MLP-based state-of-the-art models (\S\ref{sec:sota-architectures}).
Finally, we quantify the training efficiency overhead coming from the best optimizers (\S\ref{sec:efficiency}).

\subsection{Experimental Setup}
\label{sec:setup}

We mostly rely on the experiment setup from \citep{gorishniy2025tabm} --- a recent tabular DL study with a comprehensive hyperparameter tuning and evaluation protocol.
To make our report self-contained, we explicitly describe most of the details in this section and in the Appendix.

\textbf{Datasets.}
We evaluate optimizers on \Ndatasets datasets spanning both classification and regression. The benchmark combines standard academic datasets from prior tabular deep learning work \citep{gorishniy2025tabm, gorishniy2024tabr} with industrial datasets from the TabReD benchmark \citep{rubachev2025tabred}. For each dataset, there is a predefined train/validation/test split. For the TabReD datasets, we use the official benchmark preprocessing and temporal splits from \citet{rubachev2025tabred}. Dataset statistics are summarized in \autoref{app:datasets}.

\textbf{Models.}
We consider multiple MLP-based model families. The first part of our evaluation compares all optimizers for a standard ReLU MLP with dropout. The second part of our evaluation covers more advanced MLP-based models, including $\mathrm{MLP}^{\dagger}$ (a plain MLP augmented with piecewise-linear embeddings for numeric features from  \citet{gorishniy2022embeddings}), and TabM (a parameter-efficient ensemble of MLPs from \citet{gorishniy2025tabm}). We consider multiple TabM variants: TabM, TabM$^{\dagger}$ (TabM with feature embeddings, similarly to $\mathrm{MLP}^{\dagger}$), and TabM$_{\mathrm{Packed}}$ (a TabM variant without weight sharing).

\textbf{Data preprocessing and training}. We use the same preprocessing for all optimizers and models on a given dataset. The details of the data preprocessing are further described in the \autoref{app:experimental-details}. For classification, models are trained with cross-entropy loss; for regression, we normalize the labels for training and use mean squared error. We do not use data augmentation or external learning-rate schedules.\footnote{Methods such as Schedule-Free AdamW are evaluated in their intended schedule-free form; we simply do not add a separate schedule on top of the optimizer.} We apply global gradient clipping with threshold $1.0$ and use a predefined dataset-specific batch size (given in the \autoref{tab:dataset-statistics}). Training uses early stopping on the validation set with patience $16$, using the downstream evaluation metric on the validation set as the stopping criterion.

\textbf{Hyperparameter tuning.}
For each dataset, model family, and optimizer, we tune hyperparameters with Optuna using the TPE sampler \citep{akiba2019optuna}. Model hyperparameters and optimizer hyperparameters are tuned jointly. Each optimizer is tuned separately: learning rates are never reused across optimizers, and optimizer-specific parameters are tuned in their own search spaces. Within a given dataset and model family, all optimizers receive the same tuning budget and the same validation protocol. The full search spaces and trial budgets are reported in \autoref{app:experimental-details}.

\textbf{Evaluation.}
After the tuning, the selected hyperparameter configuration is retrained from scratch and evaluated over 10 random seeds. For classification, the metric is accuracy, except for the binary TabReD datasets where we follow the original paper and report ROC-AUC. For regression, we report RMSE.

\textbf{Aggregating results across datasets.}
To present results, we use three kinds of metrics.
\textit{First}, we compute performance ranks as described in \autoref{app:experimental-details}.
\textit{Second}, we compute $\Delta_{\mathrm{score}}$ --- a metric showing relative improvements over a given baseline. To compute $\Delta_{\mathrm{score}}$, we first convert regression RMSE to $R^2$, so that higher values indicate better performance for both classification and regression. Then, we compute a unified relative score as follows:
\[
 \Delta_{\mathrm{score}} \coloneq 100 \times \left(\frac{\mathrm{score}}{\mathrm{score}_{\mathrm{MLP[AdamW]}}} - 1\right)\%
\]
Positive values of $\Delta_{\mathrm{score}}$ indicate improvement over the plain MLP trained with AdamW.
\textit{Third}, we compute win/tie/loss counts against AdamW.
A win or loss is defined by Welch's $t$-test on the 10 seed-level test scores at significance level $\alpha = 0.05$; otherwise, the comparison is counted as a tie.

\subsection{MLP Benchmark}
\label{sec:main-benchmark}

We start with the plain MLP, where optimizer effects are easiest to interpret without additional architectural components. We aim to cover a representative set of practical alternatives to AdamW for tabular MLPs.

We consider \Noptimizers methods. As reference baselines, we include AdamW \citep{loshchilov2018decoupled} and SGD with momentum. Among Adam-family variants, we evaluate NAdamW \citep{dozat2016incorporating}, RAdam \citep{liu2019radam}, ADOPT \citep{taniguchi2024adopt}, Adan \citep{xie2024adan}, AdaBelief \citep{zhuang2020adabelief}, Cautious AdamW \citep{liang2026cautious}, and AdEMAMix \citep{pagliardini2025the}. We also include the sign-based methods Lion \citep{chen2023symbolic} and Signum \citep{bernstein2018signsgd}. Beyond these, we evaluate SOAP \citep{vyas2025soap} and Muon \citep{jordan2024muon}.
% We also include two optimizer-adjacent techniques in Schedule-Free \citep{defazio2024road} and Exponential Moving Averaging (EMA) of weights, we use both with AdamW as the base optimizer.
We also include Schedule-Free AdamW \citep{defazio2024road} and AdamW with Exponential Moving Averaging (EMA) of model weights.
Further implementation details like hyperparameter tuning spaces are provided in the \autoref{app:experimental-details}.

\begin{figure}[t]
  \centering
  \includegraphics[width=\linewidth]{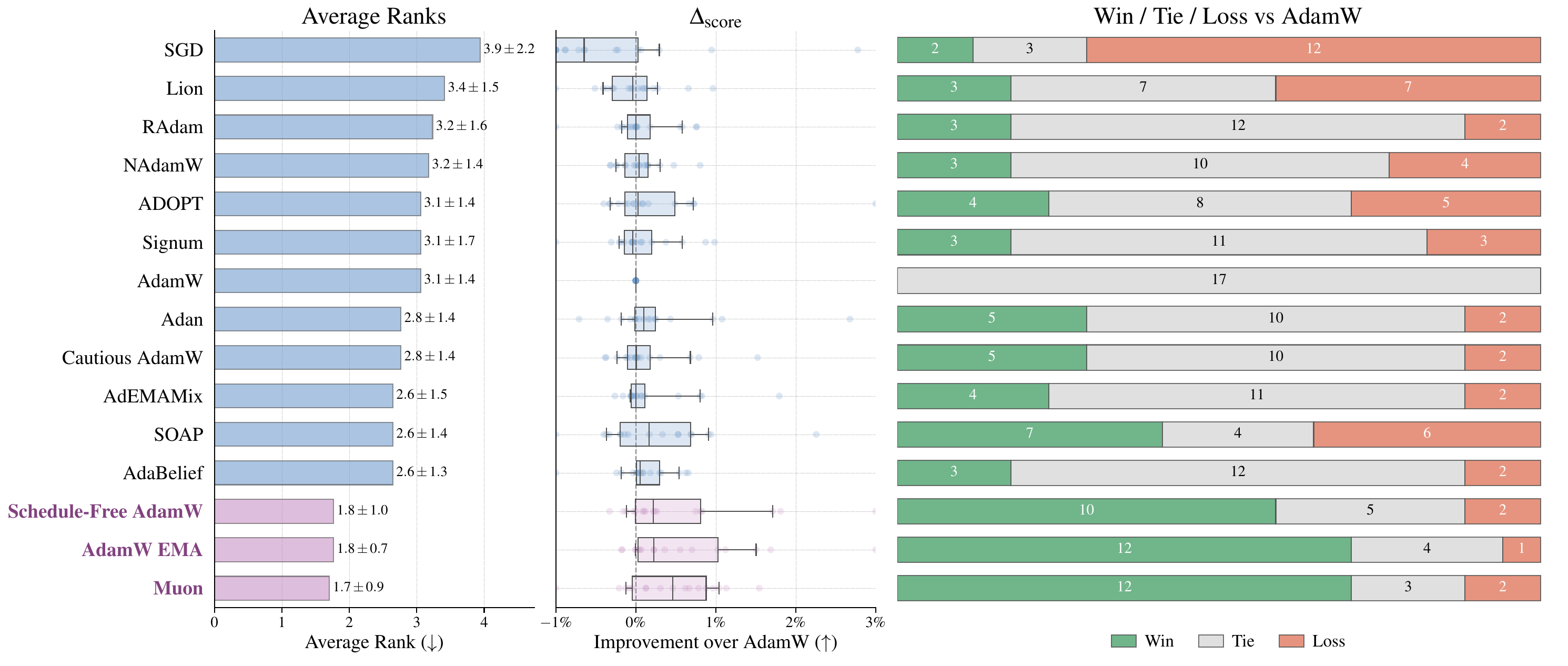}
  \caption{%
    Comparison of optimization methods for vanilla MLP across \Ndatasets datasets.
    \textbf{Left:} mean rank across datasets (lower is better).
    \textbf{Middle:} relative improvement over AdamW (higher is better). The box plots describe the percentiles of the jitter points: the boxes describe the 25th, 50th, and 75th percentiles, and the whiskers describe the 10th and 90th percentiles. Outliers are clipped.
    \textbf{Right:} pairwise win/tie/loss counts against AdamW. \textbf{\textcolor{darkpurple}{Muon, Schedule-Free AdamW, and AdamW with EMA}} consistently outperform the AdamW baseline (highlighted in bold and purple in the figure).%
  }
  \label{fig:optimizer-comparison}
\end{figure}

\textbf{Results.} \autoref{fig:optimizer-comparison} summarizes the results across the \Ndatasets datasets. Most methods remain close to AdamW and do not provide a reliable improvement in our setup. The clearest positive results come from Muon, Schedule-Free, and AdamW with EMA. These methods outperform the AdamW baseline on more than half of the datasets and yield measurable improvements in predictive performance with limited downside on individual datasets. Note that the Schedule-Free learning and EMA are conceptually related, as per discussion in recent work \citep{morwani2025connections,song2025through,zhang2025how}. Additionally, we find that combining Muon with EMA can be beneficial on some datasets, but ultimately the vanilla Muon looks like a more reliable go-to option; see \autoref{app:additonal-experiments} for details.

Based on the above results, we move on with Muon as the overall best optimizer, and AdamW with EMA as a competitive simple baseline, to the second part of our benchmark, where we focus on advanced tabular MLP-based models.

\subsection{Muon and EMA for SoTA MLP-based Models}
\label{sec:sota-architectures}

Based on the MLP benchmark above, we evaluate Muon and AdamW with EMA on stronger MLP-based model variants. Specifically, we consider $\mathrm{MLP}^{\dagger}$ with piecewise-linear embeddings for numerical features \citep{gorishniy2022embeddings}, and three TabM variants \citep{gorishniy2025tabm}: TabM, TabM$^{\dagger}$, and TabM$_{\mathrm{Packed}}$. For reference we also include the plain MLP from the previous subsection.

\textbf{Results.} \autoref{tab:optimizer-transfer} summarizes the results. Muon remains the strongest and most consistent optimizer across architectures: it improves over AdamW on every model family we evaluate. EMA gains are smaller and less uniform on stronger architectures than on the plain MLP, as indicated by the win/tie/loss counts in \autoref{tab:optimizer-transfer}.

\begin{table*}[h]
\centering
\caption{
Transfer of optimizer gains across MLP-based architectures. $\Delta_{\mathrm{score}}$
denotes the mean relative score improvement with respect to the MLP[AdamW]
baseline. The AdamW column thus shows the gain from the architecture alone. For
AdamW EMA and Muon we report the combined architecture+optimizer gain, and in
parentheses the additional gain over AdamW within the same architecture. W/T/L
compares each method to AdamW \textit{within the same model}. The last row
contains summary, which is mean for the $\Delta_{\mathrm{score}}$ columns and sum for
the W/T/L columns. Muon is the most consistent in terms of reliability of
improvement across architecture variants. EMA gains are less uniform across
architectures and datasets.}
\label{tab:optimizer-transfer}
\small
\setlength{\tabcolsep}{5pt}
\renewcommand{\arraystretch}{1.08}
\begin{tabular}{@{}lccc@{\hspace{16pt}}cc@{}}
\toprule
& \multicolumn{3}{c}{$\Delta_{\mathrm{score}}$ over MLP[AdamW]} & \multicolumn{2}{c}{W/T/L vs \textit{Architecture}[AdamW]} \\
\cmidrule(lr){2-4} \cmidrule(lr){5-6}
Model & AdamW & AdamW {\scriptsize EMA} & Muon & AdamW {\scriptsize EMA} & Muon \\
\midrule
MLP & 0.00 & 0.66 {\scriptsize (\textcolor{OliveGreen}{+\textbf{0.66}})} & 0.32 {\scriptsize (\textcolor{OliveGreen}{+\textbf{0.32}})} & 12/4/1 & 12/3/2 \\
MLP$^\dagger$ & 1.12 & 1.42 {\scriptsize (\textcolor{OliveGreen}{+\textbf{0.30}})} & 1.56 {\scriptsize (\textcolor{OliveGreen}{+\textbf{0.44}})} & 8/6/3 & 10/7/0 \\
TabM$_{\text{Packed}}$ & 1.24 & 1.31 {\scriptsize (\textcolor{OliveGreen}{+\textbf{0.08}})} & 1.45 {\scriptsize (\textcolor{OliveGreen}{+\textbf{0.21}})} & 5/8/4 & 12/3/2 \\
TabM & 1.50 & 1.75 {\scriptsize (\textcolor{OliveGreen}{+\textbf{0.26}})} & 1.71 {\scriptsize (\textcolor{OliveGreen}{+\textbf{0.21}})} & 8/7/2 & 11/5/1 \\
TabM$^\dagger$ & 2.17 & 2.21 {\scriptsize (\textcolor{OliveGreen}{+\textbf{0.04}})} & 2.57 {\scriptsize (\textcolor{OliveGreen}{+\textbf{0.40}})} & 2/10/5 & 10/7/0 \\
\midrule
\textbf{Summary:} & \textbf{1.20} & \textbf{1.47} {\scriptsize (\textbf{\textcolor{OliveGreen}{+\textbf{0.27}}})} & \textbf{1.52} {\scriptsize (\textbf{\textcolor{OliveGreen}{+\textbf{0.32}}})} & \textbf{35/35/15} & \textbf{55/25/5} \\
\bottomrule
\end{tabular}
\end{table*}

\subsection{Training Efficiency Overhead}
\label{sec:efficiency}

In \autoref{tab:efficiency}, we quantify the training efficiency overhead coming from using Muon and AdamW with EMA instead of the vanilla AdamW.
Notably, the picture is different for simpler MLPs and for the more powerful TabM.
First, for TabM, the relative efficiency overhead associated with Muon is generally less prounounced.
Second, for MLP, AdamW with EMA is noticeably more efficient than Muon, which is not the case for TabM.
% Thus, Muon seems to be a better choice for TabM than AdamW with EMA from both task performance and efficiency perspectives.

\begin{table*}[h]
\centering
\caption{
    Task performance and training efficiency overhead across multiple model families and optimizers. EMA denotes AdamW with EMA.
    The top row reports the mean improvement relative to MLP trained with AdamW ($\Delta_{\mathrm{Score}}$ is defined in \autoref{sec:setup}).
    The bottom row reports the hyperparameter tuning time overhead relative to AdamW.
    For both metrics, their mean values across all datasets are reported.
    }
\label{tab:efficiency}
\small
\setlength{\tabcolsep}{4.5pt}
\renewcommand{\arraystretch}{1.08}
\begin{tabular}{@{}l*{10}{c}@{}}
\toprule
& \multicolumn{2}{c}{MLP} & \multicolumn{2}{c}{MLP$^\dagger$} & \multicolumn{2}{c}{TabM} & \multicolumn{2}{c}{TabM$_{\text{Packed}}$} & \multicolumn{2}{c}{TabM$^\dagger$} \\
\cmidrule(lr){2-3} \cmidrule(lr){4-5} \cmidrule(lr){6-7} \cmidrule(lr){8-9} \cmidrule(lr){10-11}
Metric & EMA & Muon & EMA & Muon & EMA & Muon & EMA & Muon & EMA & Muon \\
\midrule
$\Delta_{\mathrm{Score}}$ over MLP[AdamW] & 0.66 & 0.32 & 1.42 & 1.56 & 1.75 & 1.71 & 1.31 & 1.45 & 2.21 & \textbf{2.57} \\
Time overhead over \textit{Model}[AdamW] & 1.22$\times$ & 2.98$\times$ & 1.16$\times$ & 2.28$\times$ & 2.21$\times$ & 1.18$\times$ & 1.74$\times$ & 1.57$\times$ & 1.39$\times$ & 1.24$\times$ \\
\bottomrule
\end{tabular}
\end{table*}

\section{Limitations}

Our study focuses specifically on MLP-based architectures for supervised learning on tabular data. We do not cover tabular foundation models, retrieval-based methods, or other non-MLP paradigms, where optimizer behavior may differ. Our results are also purely empirical: understanding why methods such as Muon help in this setting is an important direction for future work. Finally, replacing AdamW with Muon results in slower training, though the overhead for more powerful models is less noticeable.

\section{Conclusion}

In this report, we revisited optimizer choice for MLP-based models in tabular deep learning by benchmarking \Noptimizers methods on \Ndatasets supervised learning tasks in a unified experiment setting.
We found Muon to perform best, and hence recommend it as a powerful modern baseline for practice and research, if the associated efficiency overhead is tolerable.
Plus, for vanilla MLPs, we highlighted approaches based on model weight averaging as notable runner-ups, which includes AdamW with EMA as a simple option, and Schedule-Free AdamW as an alternative.

\bibliography{bibliography}
\bibliographystyle{tmlr}

%%%%%%%%%%%%%%%%%%%%%%%%%%%%%%%%%%%%%%%%%%%%%%%%%%%%%%%%%%%%%%%%%%%%%%%%%%%%%%%
%%%%%%%%%%%%%%%%%%%%%%%%%%%%%%%%%%%%%%%%%%%%%%%%%%%%%%%%%%%%%%%%%%%%%%%%%%%%%%%
% APPENDIX
%%%%%%%%%%%%%%%%%%%%%%%%%%%%%%%%%%%%%%%%%%%%%%%%%%%%%%%%%%%%%%%%%%%%%%%%%%%%%%%
%%%%%%%%%%%%%%%%%%%%%%%%%%%%%%%%%%%%%%%%%%%%%%%%%%%%%%%%%%%%%%%%%%%%%%%%%%%%%%%
\newpage
\appendix
\onecolumn
\section{Additional Experiments}
\label{app:additonal-experiments}

\subsection{Combining Muon with EMA}

Results of applying both EMA and Muon compared to the plain Muon are in the \autoref{tab:muon-vs-muon-ema}. We can see that EMA provides only a marginal gain in relative score and no gain in the overall amount of wins over AdamW.

\begin{table}[h]
\centering
\caption{Comparison of Muon with Muon EMA on MLP.
$\Delta_{\mathrm{score}}$ is the mean relative unified score (\%)
with respect to AdamW; the parenthesized value shows the improvement
over AdamW.  W/T/L counts are based on Welch's $t$-test ($\alpha=0.05$)
across 17 datasets. Muon EMA is slightly better in the average improvement metric, but almost the same in the W/T/L.}
\label{tab:muon-vs-muon-ema}
\small
\setlength{\tabcolsep}{5pt}
\renewcommand{\arraystretch}{1.08}
\begin{tabular}{@{}lcc@{}}
\toprule
& $\Delta_{\mathrm{score}}$ vs AdamW & W/T/L vs AdamW \\
\midrule
AdamW & 0.00 & --- \\
Muon & 0.32 {\scriptsize (\textcolor{OliveGreen}{+\textbf{0.32}})} & 12/3/2 \\
Muon EMA & 0.42 {\scriptsize (\textcolor{OliveGreen}{+\textbf{0.42}})} & 11/5/1 \\
\bottomrule
\end{tabular}
\end{table}

\section{Dataset Statistics}
\label{app:datasets}

\autoref{tab:dataset-statistics} summarizes the datasets used in our experiments. We use a subset of standard academic datastes from \citep{gorishniy2025tabm, gorishniy2024tabr}, excluding the ones from \citep{grinsztajn2022why}. We also use the TabReD benchmark~\citep{rubachev2025tabred} featuring industrial datasets with temporal train-test splits. Together, these cover a diverse range of domains, sizes, and task types.

\begin{table}[h]
\centering
\caption{Extended properties of datasets used in our study. Here, ``\# Train'', ``\# Val'', ``\# Test'' denotes the size of the corresponding dataset split; similarly, ``\# Num'', ``\# Bin'', ``\# Cat'' denotes the number of numerical, binary, and categorical features, respectively.}

\label{tab:dataset-statistics}
\setlength{\tabcolsep}{4pt} 
\fontsize{9}{10}\selectfont 
\scalebox{0.85}{
\begin{tabular}{lccccccccc}
\toprule 
Name & Source & \# Train & \# Val & \# Test & \# Num & \# Bin & \# Cat & Task Type &Batch Size\\
\midrule
Churn Modelling &  \multirow{8}{*}{\begin{tabular}[c]{@{}c@{}}\textsc{TabM}\\ {\citep{gorishniy2025tabm}}\end{tabular}}  & $6\,400$ & $1\,600$ & $2\,000$ & $7$ & $3$ & $1$ & Binclass & 128 \\
California Housing & & $13\,209$ & $3\,303$ & $4\,128$ & $8$ & $0$ & $0$ & Regression & 256 \\
House 16H & & $14\,581$ & $3\,646$ & $4\,557$ & $16$ & $0$ & $0$ & Regression & 256 \\
Adult & & $26\,048$ & $6\,513$ & $16\,281$ & $6$ & $1$ & $8$ & Binclass & 256 \\
Diamond & & $34\,521$ & $8\,631$ & $10\,788$ & $6$ & $0$ & $3$ & Regression & 512 \\
Otto Group Products & & $39\,601$ & $9\,901$ & $12\,376$ & $93$ & $0$ & $0$ & Multiclass & 512 \\
Higgs Small & & $62\,751$ & $15\,688$ & $19\,610$ & $28$ & $0$ & $0$ & Binclass & 512 \\
Black Friday & & $106\,764$ & $26\,692$ & $33\,365$ & $4$ & $1$ & $4$ & Regression & 512 \\
Microsoft & & $723\,412$ & $235\,259$ & $241\,521$ & $131$ & $5$ & $0$ & Regression & 1024 \\
\midrule
Sberbank Housing & \multirow{8}{*}{\begin{tabular}[c]{@{}c@{}}\textsc{TabReD}\\ {\citep{rubachev2025tabred}}\end{tabular}} & 18 847 & 4 827 & 4 647 & 365 & 17 & 10 & Regression & 256 \\
Ecom Offers & & $109\,341$ & $24\,261$ & $26\,455$ & $113$ & $6$ & $0$ & Binclass & 1024 \\
Maps Routing & & $160\,019$ & $59\,975$ & $59\,951$ & $984$ & $0$ & $2$ & Regression & 1024 \\
Homesite Insurance & & $224\,320$ & $20\,138$ & $16\,295$ & $253$ & $23$ & $23$ & Binclass & 1024 \\
Cooking Time & & $227\,087$ & $51\,251$ & $41\,648$ & $186$ & $3$ & $3$ & Regression & 1024 \\
Homecredit Default & & $267\,645$ & $58\,018$ & $56\,001$ & $612$ & $2$ & $82$ & Binclass & 1024 \\
Delivery ETA & & $279\,415$ & $34\,174$ & $36\,927$ & $221$ & $1$ & $1$ & Regression & 1024 \\
Weather & & $340\,596$ & $42\,359$ & $40\,840$ & $100$ & $3$ & $0$ & Regression & 1024 \\
\bottomrule
\end{tabular}
}
\end{table}

%%% Local Variables:
%%% mode: LaTeX
%%% TeX-master: "../main"
%%% End:

\section{Additional Experimental Protocol Details}
\label{app:experimental-details}

We mostly follow the experiment setup from \citep{gorishniy2025tabm}. As such, relevant parts of the text below are copied from \citep{gorishniy2025tabm}.

The source code for reproducing the benchmark is available at \url{https://github.com/yandex-research/tabular-dl-optimizers}. Some experimental details are best looked up in the source-code directly. Below we provide details not covered in the main text with pointers to the code.

\textbf{Additional Data Preprocessing Details.}
For each dataset, for all optimizers, the same preprocessing was used for fair comparison.
For numerical features, by default, we used a slightly modified version of the quantile normalization from the Scikit-learn package \citep{pedregosa2011scikit} (see the source-code, the \texttt{lib/data.py} file). There are exceptions in Otto and three TabReD datasets (Cooking Time, Delivery ETA, and Maps Routing), on Otto normalization is not used and the three TabReD datasets are already normalized.
For categorical features, we always used one-hot encoding.

\textbf{Hyperparameter tuning Details.}
We use 100 iterations with minor exceptions for larger datasets and more computationally expensive methods, these differences together with model hyperparameters are outlined in this subsection in \autoref{A:tab:mlp-space} through \autoref{A:tab:tabm-packed-space})

\textbf{Mean Rank Computation Details.} Our method of computing ranks used in \autoref{fig:optimizer-comparison} does not count small improvements as wins, which leads to a reduced range of ranks compared to a strict ordering. Intuitively, the resulting ranks can be interpreted as performance tiers.

On a given dataset, the performance of a method $A$ is summarized by its mean score $\mu_A$ and standard deviation $\sigma_A$, computed over multiple random seeds. Assuming that higher is better, we define method $A$ to be better than method $B$ if
\[
\mu_A - \sigma_A > \mu_B.
\]
In other words, a method is considered better only if its mean score is sufficiently higher, with a margin larger than one standard deviation.

When ranking multiple methods on a dataset, we first sort them in decreasing order of mean score. Starting from the top method, which is assigned rank 1, we assign the same rank to all methods that are not worse than the current reference according to the rule above. The first method that is worse than the current reference is assigned the next rank and becomes the new reference. We continue this process until all methods are ranked. Ranks are computed independently for each dataset.

The code used for the rank computation is in \texttt{lib/results.py} in the \texttt{\_compute\_ranks\_impl\_(...)} function.

\subsection{MLP}
\autoref{A:tab:mlp-space} provides hyperparameter tuning space for MLP.

\begin{table}[h!]
\centering
\caption{The hyperparameter tuning space for MLP.}
{\renewcommand{\arraystretch}{1.2}
\begin{tabular}{lll}
    \toprule
    Parameter           & Distribution \\
    \midrule
    \# layers           & $\mathrm{UniformInt}[1,6]$ \\
    Width (hidden size) & $\mathrm{UniformInt}[64,1024]$ with step 16\\
    Dropout rate        & $\{0.0, \mathrm{Uniform}[0.0,0.5]\}$ \\
    Weight decay        & $\{0, \mathrm{LogUniform}[0.005, 5.0]\}$ \\
    \midrule
    \# Tuning iterations & 100 \\
    \bottomrule
\end{tabular}}
\label{A:tab:mlp-space}
\end{table}

\subsection{MLP$^\dagger$}
\autoref{A:tab:mlp-emb-space} provides hyperparameter tuning space for MLP$^\dagger$.

\begin{table}[h!]
\centering
\caption{The hyperparameter tuning space for $\mathrm{MLP}^{\dagger}$.}
{\renewcommand{\arraystretch}{1.2}
\begin{tabular}{lll}
    \toprule
    Parameter           & Distribution \\
    \midrule
    \# layers           & $\mathrm{UniformInt}[1,5]$ \\
    Width (hidden size) & $\mathrm{UniformInt}[64,1024]$ with step 16 \\
    Dropout rate        & $\{0.0, \mathrm{Uniform}[0.0,0.5]\}$ \\
    Weight decay        & $\{0, \mathrm{LogUniform}[0.001, 1.0]\}$ \\
    \midrule
    d\_embedding      & $\mathrm{UniformInt}[8,32]$ with step 4 \\
    n\_bins      & $\mathrm{UniformInt}[2,128]$ \\
    \# Tuning iterations & 100 \\
    \bottomrule
\end{tabular}}
\label{A:tab:mlp-emb-space}
\end{table}

\subsection{TabM}
\autoref{A:tab:tabm-space} provides hyperparameter tuning space for TabM. \autoref{A:tab:tabm-packed-space} provides hyperparameter tuning space for TabM-packed.

\begin{table}[h!]
\centering
\caption{The hyperparameter tuning space for TabM. Here, (B) = \{Microsoft and TabRed datasets\} and (A) contains all other datasets.}
{\renewcommand{\arraystretch}{1.2}
\begin{tabular}{lll}
    \toprule
    Parameter           & Distribution or Value\\
    \midrule
    $k$                   & $16$ \\
    \# layers           & $\mathrm{UniformInt}[1,5]$ \\
    Width (hidden size) & $\mathrm{UniformInt}[64,1024]$ with step 16\\
    Dropout rate        & $\{0.0, \mathrm{Uniform}[0.0,0.5]\}$ \\
    Weight decay        & $\{0, \mathrm{LogUniform}[0.001, 1.0]\}$ \\
    \midrule
    \# Tuning iterations & (A) 100  (B) 50 \\
    \bottomrule
\end{tabular}}
\label{A:tab:tabm-space}
\end{table}

\begin{table}[h!]
\centering
\caption{The hyperparameter tuning space for TabM-packed. Here, (B) = \{Microsoft and TabRed datasets\} and (A) contains all other datasets.}
{\renewcommand{\arraystretch}{1.2}
\begin{tabular}{lll}
    \toprule
    Parameter           & Distribution or Value\\
    \midrule
    $k$                   & $16$ \\
    \# layers           & $\mathrm{UniformInt}[1,5]$ \\
    Width (hidden size) & $\mathrm{UniformInt}[64,1024]$ with step 16\\
    Dropout rate        & $\{0.0, \mathrm{Uniform}[0.0,0.5]\}$ \\
    Weight decay        & $\{0, \mathrm{LogUniform}[0.005, 5.0]\}$ \\
    \midrule
    \# Tuning iterations & (A) 100  (B) 50 \\
    \bottomrule
\end{tabular}}
\label{A:tab:tabm-packed-space}
\end{table}

\subsection{Optimizers}

We report hyperparameter tuning spaces related to optimizers, this spaces are added to the model spaces and tuned jointly using model-specific number of tuning iterations. All the optimizer-specific hyperparameters are in \autoref{A:tab:optimizer-spaces}.

\section{Per-Dataset Results}
\label{app:per-dataset}

This section provides detailed per-dataset results for the main experiments. All metrics are averaged over 10 random seeds; standard deviations are shown in parentheses. Bold values indicate the best result within each comparison group. Results for all methods on all datasets are in the \autoref{tab:appendix-all-results}.

\newpage

\begin{table*}[h]
\centering
\caption{Optimizer hyperparameter tuning spaces for the MLP benchmark.
Spaces are added to the model space (\autoref{A:tab:mlp-space})
and tuned jointly (100 iterations).}
\label{A:tab:optimizer-spaces}
\small
\setlength{\tabcolsep}{4pt}
\renewcommand{\arraystretch}{1.12}

\begin{minipage}[t]{0.48\textwidth}
\vspace{0pt}
\begin{tabular}{@{}ll@{}}
\toprule
Parameter & Distribution / Value \\
\midrule
\multicolumn{2}{l}{\textbf{AdamW}} \\
\quad Learning Rate & $\mathrm{LogUniform}[3 \!\times\! 10^{-5},\,0.001]$ \\
\quad Weight Decay & $\mathrm{LogUniform}[0.005,\,5]$ \\
\midrule
\multicolumn{2}{l}{\textbf{AdamW\ EMA}} \\
\quad EMA decay & $\mathrm{LogUniform}[0.9,\,0.999]$ \\
\quad Learning Rate & $\mathrm{LogUniform}[3 \!\times\! 10^{-5},\,0.001]$ \\
\quad Weight Decay & $\mathrm{LogUniform}[0.005,\,5]$ \\
\midrule
\multicolumn{2}{l}{\textbf{NAdamW}} \\
\quad Learning Rate & $\mathrm{LogUniform}[3 \!\times\! 10^{-5},\,0.001]$ \\
\quad Weight Decay & $\mathrm{LogUniform}[0.005,\,5]$ \\
\midrule
\multicolumn{2}{l}{\textbf{RAdam}} \\
\quad Learning Rate & $\mathrm{LogUniform}[3 \!\times\! 10^{-5},\,0.001]$ \\
\quad Weight Decay & $\mathrm{LogUniform}[0.005,\,5]$ \\
\midrule
\multicolumn{2}{l}{\textbf{AdaBelief}} \\
\quad Learning Rate & $\mathrm{LogUniform}[3 \!\times\! 10^{-5},\,0.001]$ \\
\quad Weight Decay & $\mathrm{LogUniform}[0.005,\,5]$ \\
\midrule
\multicolumn{2}{l}{\textbf{ADOPT}} \\
\quad Learning Rate & $\mathrm{LogUniform}[3 \!\times\! 10^{-5},\,0.001]$ \\
\quad Weight Decay & $\mathrm{LogUniform}[0.005,\,5]$ \\
\midrule
\multicolumn{2}{l}{\textbf{Adan}} \\
\quad Learning Rate & $\mathrm{LogUniform}[0.0001,\,0.01]$ \\
\quad Weight Decay & $\mathrm{LogUniform}[0.005,\,5]$ \\
\midrule
\multicolumn{2}{l}{\textbf{AdEMAMix}} \\
\quad Learning Rate & $\mathrm{LogUniform}[3 \!\times\! 10^{-5},\,0.001]$ \\
\quad $\alpha$ & $\mathrm{Uniform}[1,\,8]$ \\
\quad Weight Decay & $\mathrm{LogUniform}[0.005,\,5]$ \\
\midrule
\multicolumn{2}{l}{\textbf{Lion}} \\
\quad Learning Rate & $\mathrm{LogUniform}[10^{-5},\,0.001]$ \\
\quad Weight Decay & $\mathrm{LogUniform}[0.005,\,5]$ \\
\bottomrule
\end{tabular}
\end{minipage}
\hfill
\begin{minipage}[t]{0.48\textwidth}
\vspace{0pt}
\begin{tabular}{@{}ll@{}}
\toprule
Parameter & Distribution / Value \\
\midrule
\multicolumn{2}{l}{\textbf{Signum}} \\
\quad Learning Rate & $\mathrm{LogUniform}[10^{-5},\,0.001]$ \\
\quad Weight Decay & $\mathrm{LogUniform}[0.005,\,5]$ \\
\midrule
\multicolumn{2}{l}{\textbf{Cautious AdamW}} \\
\quad Learning Rate & $\mathrm{LogUniform}[3 \!\times\! 10^{-5},\,0.001]$ \\
\quad Weight Decay & $\mathrm{LogUniform}[0.005,\,5]$ \\
\midrule
\multicolumn{2}{l}{\textbf{SOAP}} \\
\quad Learning Rate & $\mathrm{LogUniform}[3 \!\times\! 10^{-5},\,0.001]$ \\
\quad Weight Decay & $\mathrm{LogUniform}[0.005,\,5]$ \\
\midrule
\multicolumn{2}{l}{\textbf{ScheduleFree}} \\
\quad Learning Rate & $\mathrm{LogUniform}[0.0001,\,0.03]$ \\
\quad Weight Decay & $\mathrm{LogUniform}[0.005,\,5]$ \\
\midrule
\multicolumn{2}{l}{\textbf{SGD}} \\
\quad Momentum & $0.9$ \\
\quad Dampening & $0.9$ \\
\quad Learning Rate & $\mathrm{LogUniform}[0.001,\,0.1]$ \\
\quad Weight Decay & $\mathrm{LogUniform}[0.005,\,5]$ \\
\midrule
\multicolumn{2}{l}{\textbf{Muon}} \\
\quad Learning Rate & $\mathrm{LogUniform}[3 \!\times\! 10^{-5},\,0.001]$ \\
\quad Weight Decay & $\mathrm{LogUniform}[0.005,\,5]$ \\
\quad $(\beta_1, \beta_2)$ & $(0.9,\,0.999)$ \\
\quad $\epsilon$ & $10^{-8}$ \\
\quad Muon LR & $\mathrm{LogUniform}[0.0001,\,0.03]$ \\
\midrule
\multicolumn{2}{l}{\textbf{Muon EMA}} \\
\quad EMA decay & $\mathrm{LogUniform}[0.9,\,0.999]$ \\
\quad $(\beta_1, \beta_2)$ & $(0.9,\,0.999)$ \\
\quad $\epsilon$ & $10^{-8}$ \\
\quad Learning Rate & $\mathrm{LogUniform}[3 \!\times\! 10^{-5},\,0.001]$ \\
\quad Weight Decay & $\mathrm{LogUniform}[0.005,\,5]$ \\
\quad Muon LR & $\mathrm{LogUniform}[0.0001,\,0.03]$ \\
\bottomrule
\end{tabular}
\end{minipage}

\end{table*}

\newpage

\newcommand{\topalign}[1]{%
\vtop{\vskip 0pt #1}}

\begin{longtable}{p{0.5\textwidth}p{0.5\textwidth}}
\caption{
Per-dataset results for all methods evaluated in this paper. For each dataset we report the mean test score $\pm$ standard deviation over 10 random seeds.
}
\label{tab:appendix-all-results}\\
\end{longtable}

\end{document}